%% file: main.tex
\documentclass{article}

\PassOptionsToPackage{numbers, sort&compress}{natbib}

    \usepackage[final]{neurips_2020}

\usepackage[utf8]{inputenc} 
\usepackage[T1]{fontenc}    
\usepackage{hyperref}       
\usepackage{url}            
\usepackage{booktabs}       
\usepackage{amsfonts}       
\usepackage{nicefrac}       
\usepackage{microtype}      

\usepackage[small]{caption}
\usepackage{graphicx}
\usepackage{amsmath}
\usepackage{amsthm}
\usepackage{amssymb}
\usepackage{mathtools}
\usepackage{algorithm}
\usepackage{algorithmic}
\usepackage{paralist}
\usepackage{latexsym}
\usepackage[title]{appendix}
\usepackage{subcaption}
\usepackage{subfiles}
\usepackage{csquotes}
\usepackage{epsfig}
\usepackage{tikz}
\usepackage[super]{nth}
\usepackage{cancel}
\usepackage{multirow}

\newcommand{\mb}[1]{\mathbf{#1}}
\newcommand{\grad}{\nabla_\theta}

\newcommand{\pt}{\pi_\theta}
\newcommand{\expt}[1]{\underset{#1}{\mathbb{E}}}
\newcommand{\replace}[2]{[{#1}_{<t}, #2, {#1}_{>t}]}
\newcommand{\Rtilde}{\widetilde{R}}
\newcommand{\pre}[1]{#1_{<t}}
\newcommand{\ind}{\mathbb{I}}

\newcommand{\Table}[1]{Table~\ref{#1}}
\newcommand{\Fig}[1]{Figure~\ref{#1}}
\newcommand{\Equation}[1]{(\ref{#1})}
\newcommand{\Section}[1]{Section~\ref{#1}}
\newcommand{\Appendix}[1]{Appendix~\ref{#1}}
\newcommand{\shortcite}[1]{\citet{#1}}

\title{TaylorGAN: Neighbor-Augmented Policy Update for Sample-Efficient Natural Language Generation}

\author{
    Chun-Hsing Lin \quad Siang-Ruei Wu \quad Hung-Yi Lee \quad Yun-Nung Chen \\
    National Taiwan University, Taipei, Taiwan \\
    \texttt{\{jsaon92, raywu0\}@gmail.com \quad hungyilee@ntu.edu.tw \quad y.v.chen@ieee.org} \\
}

\begin{document}

\maketitle

\begin{abstract}
Score function-based natural language generation (NLG) approaches such as REINFORCE, in general, suffer from low sample efficiency and training instability problems.
This is mainly due to the non-differentiable nature of the discrete space sampling and thus these methods have to treat the discriminator as a black box and ignore the gradient information.
To improve the sample efficiency and reduce the variance of REINFORCE, we propose a novel approach, TaylorGAN, which augments the gradient estimation by off-policy update and the first-order Taylor expansion.
This approach enables us to train NLG models from scratch with smaller batch size --- without maximum likelihood pre-training, and outperforms existing GAN-based methods on multiple metrics of quality and diversity.\footnote{The source code and data are available at \url{https://github.com/MiuLab/TaylorGAN/}}
\end{abstract}

\section{Introduction}
\input{sections/01-introduction.tex}

\section{Background}\label{sec:related-work}
\input{sections/02-related_work}

\section{Proposed Method}\label{sec:proposed-method}
\input{sections/03-proposed_method.tex}

\section{Experiments}
\input{sections/04-experiments.tex}
\section{Results \& Discussion}
\input{sections/05-results}

\section{Conclusion and Future Work}
\input{sections/06-conclusion}

\section*{Broader Impact}
\input{sections/broader_impact}

\begin{ack}
We would like to thank all reviewers for their insightful comments;
Alvin Chiang, Chien-Fu Lin, Chi-Liang Liu, Po-Hsien Chu, Yi-En Tsai and Chung-Yang (Ric) Huang for thoughtful discussions; Shu-Wen Yang and Huan Lin for technical supports.
We are grateful to the National Center for High-performance Computing for computing resources.
This work was financially supported from the Young Scholar Fellowship Program by Ministry of Science and Technology (MOST) in Taiwan,
under Grant 109-2636-E-002-026 and Grant 109-2636-E-002-027.

\end{ack}

\bibliographystyle{named}
\bibliography{references}

\appendix
\input{sections/appendix}

\end{document}

%% file: sections/01-introduction.tex
Generative adversarial networks (GAN) \citep{NIPS2014_5423} have advanced many applications such as image generation and unsupervised style transfer \citep{2015arXiv151106434R,karras2019style}. Unsurprisingly, much effort has been devoted to adopting the GAN framework for unsupervised text generation \citep{MaliGAN2017,Yu2017SeqGANSG,zhang2017adversarial,Guo2018LongTG,fedus2018maskgan,nie2018relgan,ScratchGAN}. 
However, in natural language generation (NLG), more challenges are concerned, such as passing discrete tokens through a non-differentiable operation, which prohibits backpropagating the gradient signal to the generator.

To address the issue about non-differentiablity, researchers and practitioners used score function-based gradient estimators such as REINFORCE to train GANs for NLG, where the discriminator is cast as a reward function for the generator. 
These methods suffer from poor sample efficiency, high variance, and credit assignment problems.
We argue that it is disadvantageous to utilize the discriminator as a simple reward function when it is known that gradient-based backpropagation is more effective for optimization.

In this paper, we propose a novel unsupervised NLG technique, TaylorGAN, where the approximated reward of sequences improves the efficiency and accuracy of the estimator.
Our contributions are 3-fold:
\begin{compactitem}
   \item This paper proposes a novel update formula for the generator to incorporate gradient information from the discriminator during training.
   \item The experiments demonstrate that TaylorGAN achieves state-of-the-art performance without maximum likelihood pre-training.
   \item Our model does not require additional variance reduction techniques used in other REINFORCE-based counterparts such as large batch size \cite{ScratchGAN}, value estimation model \cite{Guo2018LongTG} and Monte-Carlo rollouts \cite{Yu2017SeqGANSG}.
\end{compactitem}

%% file: sections/02-related_work.tex

GAN is an innovative approach of generative modeling. Instead of learning a probabilistic model via maximum likelihood estimation (MLE), GAN is a two-player minimax game, in which the generator $G_{\theta}$ aims at mapping random noises $z$ to realistic samples and the discriminator $D_{\phi}$ focuses on classifying whether a sample is from real data $p_{data}$ or from the generator.
\begin{align}
    \min_\theta \max_\phi \expt{x \sim p_{data}} [\log D_\phi (x)] + \expt{z \sim p_z}[\log(1 -  D_\phi(G_\theta(z)))]
\end{align}

In the standard architecture of GAN, the generator’s output $G_\theta(z)$ is directly connected as the input to the discriminator in a fully differentiable manner, which means that the gradients of the objective $\grad \log(1 -  D_\phi(G_\theta(z))) $ can be directly backpropagated to the generator's parameters $\theta$. However, in NLG, data are defined in a discrete domain, which is essentially \emph{non-differentiable}. In order to avoid the intractability of gradients, text GANs proposed various approaches for estimating the gradients.

\paragraph{Continuous Relaxation} \label{paragraph:continuous}
Continuous relaxation approaches such as Gumbel-Softmax~\cite{jang2016categorical} approximate discrete data in terms of continuous variables such as outputs of a softmax function. While it allows us to ignore the non-differentiable discrete sampling~\cite{Kusner2016GANSFS,nie2018relgan}, several issues may be occurred.

First of all, the discriminator only needs to spot the difference between the discrete real data and the continuous softmax outputs, so the generator may learn to produce extremely ``spiky'' predictions.
Therefore, this training procedure creates a major inconsistency.
Specifically, during testing, the generator has to sample a sequence of discrete tokens from the distribution, whereas during training, it is only trained to generate a feasible expectation, which may result in non-realistic texts.

\paragraph{Score Function-Based Gradient Estimator}
The score function-based estimator \cite{fu2006gradient,glynn1990likelihood}, also known as REINFORCE \cite{williams1992simple}, is a common solution for addressing the non-differentiable issue mentioned above.
By directly parametrizing the probability mass function (PMF) of the sample $x$ as $p_\theta(x)$ and using the identity $\grad p_\theta = p_\theta \grad \log p_\theta$, the gradient of the expectation of a function $f$ can be written as
\begin{align}
    \grad \expt{x \sim p_{\theta}} [f(x)] = 
    \expt{x \sim p_{\theta}} [f(x)\grad \log p_{\theta}(x)].
    \label{eq:reinforce}
\end{align}

Note that $f$ can be \emph{non-differentiable}. 
Although REINFORCE is an \emph{unbiased} estimator, it still has a number of disadvantages such as \emph{high variance}, \emph{low sample efficiency} and \emph{credit assignment} problems. Therefore, a lot of efforts have been devoted to reducing the variance including providing per-word  rewards~\cite{Yu2017SeqGANSG,MaliGAN2017,fedus2018maskgan,ScratchGAN} and other methods\cite{Muprop2016,RELAX2017}.

\paragraph{Straight-Through Gradient Estimator} 
Another heuristic approach is to utilize a straight-through gradient estimator \cite{Bengio2013EstimatingOP}. The basic idea is to treat the discrete operation as if it had been a differentiable proxy during the backward pass. Specifically, using the straight-through estimator with PMF as the proxy of stochastic categorical one-hot vector \cite{jang2016categorical}, the gradient of a function $f$ with respect to parameter $\theta$ can be written as
\begin{align}
    \hat{g}^{\text{ST}}_{\theta} &\coloneqq \expt{x \sim p_\theta}[\nabla_{\mb{h}} f(\mb{h}) \big|_{\mb{h} = \text{one-hot}(x)}] \cdot \grad \mb{p}_\theta, \label{eq:st} \\
    \mb{p}_\theta &\coloneqq [p_\theta(\text{``apple''}), \dots, p_\theta(\text{``zebra''})],
\end{align}
where $\mb{p}_\theta$ is a vector of probabilities of all categories.
The deterministic and dense $\grad \mb{p}_\theta$ term considering all categories results in a \emph{biased} but \emph{low-variance} estimator~\cite{Muprop2016}.

%% file: sections/03-proposed_method.tex
\input{figures/taylor_gan}
In a general setting for probabilistic NLG, a discrete token sequence $\mb{x} \in V^T$ is generated through an auto-regressive process, where $V$ is the vocabulary set and $T$ is the length of the sentence. At each step $t$, the generator samples a new token $x_t$ from a soft policy $\pt$ given the prefix $\pre{\mb{x}}$, where the probability of generating a sequence $\mb{x}$ is:
\begin{align}
    p_\theta(\mb{x}) = \pt(\mb{x}) = \prod_{t=1}^T \pt(x_t\mid\pre{\mb{x}}). \label{eq:auto-regressive}
\end{align}

In the context of GANs with \emph{score function-based approaches}, $\mb{x}$ is fed to the discriminator for its reward $R(\mb{x})$ after all tokens are generated.
The goal of the generator is to maximize the expected reward; therefore $\pt$ is updated by $\grad \expt{\mb{x} \sim \pt}[R(\mb{x})]$.

This approach has been adopted to train NLG models with better global consistency than using MLE \cite{ScratchGAN}, but is only feasible with a large batch size that reduces the variance of REINFORCE.
For a more practical and efficient score function-based method, we propose TaylorGAN, which consists of three key components: \textbf{Taylor estimator} for variance reduction, \textbf{discriminator constraint} for learning reliable rewards and \textbf{entropy maximization} for avoiding mode dropping.

\subsection{Taylor Estimator}
The discriminator is treated as a black box reward provider in the previous score function-based approaches due to non-differentiability of the discrete input space. Ignoring the derivative of $R$, the generator can only learn through the statistics of random guesses. Noisy and inaccurate updates may happen especially when samples are insufficient as demonstrated in \Fig{fig:reinforce}. The sensitivity to samples results in \emph{high variance} and \emph{low sample efficiency} problems. Moreover, it is hard to assign proper credits to the values of each generated $x_t$ since $R$ is provided on per-sequence level. Our Taylor estimator mitigates these problems by taking into account neighboring sentences (illustrated in \Fig{fig:taylor_hamming}), whose rewards can be efficiently approximated with the gradient information from the discriminator.

\subsubsection{Augmenting Samples by Taylor Expansion}
Usually, the reward $R(\mb{x})$ of discrete sequence $\mb{x}$ is obtained with an embedding operation $E: V^T \to \mathbb{R}^{d_E\times T}$ and a differentiable function $R_E: \mathbb{R}^{d_E\times T} \to \mathbb{R}$ on the embedding space:
\begin{align}
    R(\mb{x}) &\coloneqq R_E(E(\mb{x})) , \\
    E(\mb{x}) &\coloneqq [\mb{e}_{x_1}, \dots, \mb{e}_{x_T}],
\end{align}
where $\mb{e}_v \in \mathbb{R}^{d_E}$ is the $d_E$-dimensional column vector representation of the token $v$.

With this particular construction, $R(\mb{y})$ can be approximated by the first-order Taylor expansion given $R(\mb{x})$ and $\nabla_{\mb{E}} R_E(\mb{E})$ computed by backpropagation in the embedding layer's output $\mb{E} \in \mathbb{R}^{d_E \times T}$, which $\mb{y}$ is any sequence $\in V^T$ and vec$(\cdot)$ denotes the vectorization operation \citep{10.2307/2958818} (also known as the flatten operation):
\begin{align}
    \Rtilde(\mb{x} \to \mb{y}) &\coloneqq R(\mb{x}) + \text{vec}\left( E(\mb{y}) - E(\mb{x})  \right) \cdot \text{vec}\left(\nabla_{\mb{E}} R_E(\mb{E}) \bigg|_{\mb{E} = E(\mb{x})} \right) \label{eq:taylor} \\
    &\ = R(\mb{y}) + \epsilon(\mb{x} \to \mb{y}), \nonumber\\
    |\epsilon(\mb{x} \to \mb{y})| &\ = \mathcal{O}(\Vert E(\mb{y}) - E(\mb{x})\Vert^2_2). \label{eq:remainder}
\end{align}
We denote $\Rtilde(\mb{x} \to \mb{y})$ to the first-order approximation of $R(\mb{y})$ around $E(\mb{x})$, and $\epsilon$ is the first-order Taylor remainder. The benefit of utilizing gradient information is shown in \Fig{fig:taylor}. With $\Rtilde$ defined on the tangent plane in the embedding space, the update direction becomes more accurate and less sensitive to the sample $\mb{x}$.

However, the approximation is only accurate in the neighborhood of $\mb{x}$, because $\epsilon$ increases with the distance from $\mb{x}$ in the embedding space.
To ensure the neighborhood relation, we can draw $\mb{y}$ used in \Equation{eq:taylor} from a joint distribution $\Gamma(\mb{x}, \mb{y})$, which will be specified in the next subsection, instead of applying the policy $\pt$. On this off-policy procedure, any expectation on  distribution $\pt$ can still be estimated by importance sampling with likelihood-ratio $I(\mb{y}) = \dfrac{\pt(\mb{y})}{\sum_{\mb{x}'} \Gamma(\mb{x}', \mb{y})}$:
\begin{align}
    \expt{\mb{y} \sim \pt}[f(\mb{y};R(\mb{y}))]
    &= \expt{(\mb{x}, \mb{y}) \sim \Gamma}[I(\mb{y}) f(\mb{y};R(\mb{y}))] \approx \expt{\mb{x} \sim \pt}[\sum_{\mb{y}} \frac{\Gamma(\mb{x},\mb{y})}{\pt(\mb{x})} I(\mb{y}) f(\mb{y};\Rtilde(\mb{x} \to \mb{y}))]. \label{eq:augment}
\end{align}

By \Equation{eq:augment}, the reward signal is augmented with neighboring \emph{target} $\mb{y}$s, which are approximated from a single \emph{proposal} $\mb{x}$; therefore, the estimation is more efficient than previous approaches. 

Combining Taylor estimation, off-policy sampling with the REINFORCE estimator in \Equation{eq:reinforce} and sampling $\mb{y}$ from different joint distribution $\Gamma_t$ of sequence pair $\mb{x},\mb{y}$ at each step $t$:
\begin{align}
    \expt{\mb{y} \sim \pt}[R(\mb{y}) \grad \log \pt(\mb{y})]
    &= \sum_{t=1}^T \expt{\mb{y} \sim \pt}[R(\mb{y})  \grad \log\pt(y_t\mid \pre{\mb{y}})] \nonumber \\
    &\approx \sum_{t=1}^T \expt{\mb{x} \sim \pt}[\sum_{\mb{y}} \frac{\Gamma_t(\mb{x},\mb{y})}{\pt(\mb{x})} I_t(\mb{y}) \Rtilde(\mb{x} \to \mb{y}) \grad \log\pt(y_t\mid \pre{\mb{y}})] \nonumber \\
    &= \expt{\mb{x} \sim \pt}[\sum_{t=1}^T \sum_{\mb{y}} \frac{\Gamma_t(\mb{x},\mb{y})}{\pt(\mb{x})} I_t(\mb{y}) \Rtilde(\mb{x} \to \mb{y}) \grad \log\pt(y_t\mid \pre{\mb{y}})].
    \label{eq:taylor-general}
\end{align}
$\Gamma_t(\mb{y}\mid \mb{x})\coloneqq \dfrac{\Gamma_t(\mb{x},\mb{y})}{\pt(\mb{x})}$ is defined as the \emph{transition density} that brings signal $\Rtilde(\mb{x}\to\mb{y})$ to reward $\pt$ for generating $y_t$ given the prefix $\mb{y}_{<t}$ when $\mb{x}$ is sampled. \emph{Credit assignment} is performed with $\Gamma_t$ and $\nabla_{\mb{E}}R_E$ in \Equation{eq:taylor} computed for every step $t$.

\subsubsection{Hamming Transition} \label{sec:hamming}

After developing the general concept of the samples augmentation, we need $\Gamma_t$ to establish the neighborhood in the embedding space, and we also enforce $\mb{y}_{<t} = \mb{x}_{<t}$ to avoid extra computation for $\pt(\cdot\mid \pre{\mb{y}})$ in \Equation{eq:taylor-general}. 
In this work, we choose the followings:
\begin{align}
    \Gamma_t(\mb{y}\mid \mb{x}) &\coloneqq \begin{cases}
         K(y_t\mid x_t) &, \pre{\mb{x}} = \mb{y}_{< t} \text{ and } \mb{x}_{>t} = \mb{y}_{>t}, \\
        0 &, \text{otherwise}.
    \end{cases}, \\
    K(u\mid v) &\coloneqq C(v) \cdot \text{exp}(-\dfrac{\Vert \mb{e}_{u} - \mb{e}_{v} \Vert^2_2}{2 \Lambda^2}), \label{eq:kernel}
\end{align}
where $K$ is a Gaussian kernel function on the embedding space, $C(v)$ is the normalization constant, and $\Lambda > 0$ is defined as the \emph{bandwidth} parameter of the transition.

Since $\mb{y}$ only differs from $\mb{x}$ by one token, the support of $\Gamma_t$ becomes the projection of \emph{unit Hamming sphere} (UHS) on axis $t$:
\begin{align}
    \text{supp}\left( \Gamma_t(\cdot\mid \mb{x}) \right) = \text{UHS}_t(\mb{x}) \coloneqq \left\{\replace{\mb{x}}{v} \mid v \in V \right\}. \label{eq:UHS}
\end{align}
Furthermore, all $\widetilde{R}$ can be easily computed on this support (detailed in \Appendix{appendix_sec:taylor-implement}).

To calculate the likelihood-ratio $I_t$, we need $\pt(\replace{\mb{x}}{v})$ for arbitrary $v$, which needs much more computation. We therefore ignore the effect of $v$ on the rest of auto-regressive sampling, 
\begin{align}
    \frac{\pt(\replace{\mb{x}}{v})}{\pt(\replace{\mb{x}}{u})}
    &= \frac{\pt(v \mid \pre{\mb{x}})}{\pt(u \mid \pre{\mb{x}})} \frac{\pt(\mb{x}_{>t}\mid \pre{\mb{x}},v)}{\pt(\mb{x}_{>t}\mid \pre{\mb{x}}, u)}
    \approx \frac{\pt(v \mid \pre{\mb{x}})}{\pt(u \mid \pre{\mb{x}})}, \label{eq:suffix}
\end{align}
then $I_t$ is simplified as:
\begin{align}
    I_t(\mb{y}) &= \frac{\pt(\mb{y})}{\sum_{\mb{x}' \in \text{UHS}_t(\mb{y})} \Gamma_t(\mb{y}\mid \mb{x}') \pt(\mb{x}')} \approx \frac{\pt(y_t\mid \pre{\mb{y}})}{\sum_{u \in V} K(y_t\mid u) \pt(u\mid \pre{\mb{y}})}.
\end{align}
The bias introduced by this simplification can be eliminated after replacing the reward with a partial evaluation function independent of $\mb{x}_{>t}$ (proved in \Appendix{appendix_sec:suffix}). Due to the additional complexity, we leave this variation for future work.

After subtracting rewards with the bias-free baseline $b$ (proved in \Appendix{appendix_sec:suffix}), the Taylor estimator with Hamming transition is constructed as follows:
\begin{align}
    \hat{g}^{\text{Taylor}}_\theta &\coloneqq \frac{1}{N} \sum_{n = 1}^N \sum_{t=1}^T \sum_{\mb{y} \in \text{UHS}_t(\mb{x}^{(n)})} \widetilde{A}_t(\mb{x}^{(n)} \to \mb{y}) \grad \log\pt(y_t\mid \mb{x}^{(n)}_{<t}), \label{eq:taylor-estimator} \\
    \widetilde{A}_t(\mb{x} \to \mb{y}) &\coloneqq \frac{K(y_t\mid x_t) \pt(y_t\mid \pre{\mb{x}})}{\sum_{u \in V} K(y_t\mid u) \pt(u\mid \pre{\mb{x}})} (\Rtilde(\mb{x} \to \mb{y}) - b), \\
    b &\coloneqq \text{Exponential-Moving-Average}(\frac{1}{N}\sum_{n=1}^N R(\mb{x}^{(n})),
\end{align}
where $N$ is batch size of proposal $\mb{x}$s and $\widetilde{A}_t$ is defined as the \emph{advantage} of replacing $x_t$ by $y_t$. 
The dense form of $\hat{g}^{\text{Taylor}}_\theta$ providing rewards of all sequences on the unit Hamming sphere (illustrated in \Fig{fig:taylor_hamming}) can also be interpreted as \emph{pseudo exploration}.

\paragraph{Bias-Variance Trade-Off} \label{paragraph:trade-off}
By adjusting bandwidth $\Lambda$ in \Equation{eq:kernel}, the Taylor estimator can interpolate between REINFORCE and Straight-Through estimators (proved in \Appendix{appendix_sec:RST}), which are considered as unbiased and low variance approaches respectively.

\begin{itemize}
  \item $\Lambda \to 0 \Rightarrow K(u\mid v) \to \ind(u = v)$ \\
  No transition or approximation is applied: $ \hat{g}^{\text{Taylor}}_\theta \to \hat{g}^{\text{REINFORCE}}_\theta$.

  \item $\Lambda \to \infty \Rightarrow K(u\mid v) \to 1~/~|V|$ \\
  Uniform transition ignoring the neighborhood requirement: $\hat{g}^{\text{Taylor}}_\theta \to \hat{g}^{\text{ST}}_\theta$.
\end{itemize}

\subsection{Discriminator Constraint}\label{sec:d-constraint}
After constructing the gradient estimator, we choose a proper reward for NLG.
We adopt one of the loss functions from~\shortcite{zhou2019lipschitz} for informative gradients:
\begin{align}
    \mathcal{L}_D &\coloneqq -\expt{\mb{x} \sim p_{data}} [\log D_\phi (\mb{x})] - \expt{\mb{x}\sim p_\theta}[\log(1 -  D_\phi(\mb{x}))] + \mathcal{L}_{\text{reg}} \label{eq:D-loss}, \\
    R(\mb{x}) &\coloneqq \log \frac{D_\phi(\mb{x})}{1 - D_\phi(\mb{x})} = \text{sigmoid}^{-1}(D_\phi(\mb{x})). \label{eq:logit_reward}
\end{align}
This reward is the sigmoid logit, whose unboundedness and absence of the final non-linear sigmoid transformation benefit our Taylor estimator.

We further bound the distance between all pairs of embedding vectors along with Lipschitz constant of $R_E$. Incorporating this constraint to the derivation by \citet{zhou2019lipschitz}, the optimal discriminator minimizing \Equation{eq:D-loss} can guide the generator towards real samples on the basis of Hamming distance:
\begin{align}
    \Vert R(\mb{x}) - R(\mb{y}) \Vert^2 &\le \Vert R_E \Vert_{Lip}^2 \cdot \sum_{t=1}^T \Vert \mb{e}_{x_t} - \mb{e}_{y_t} \Vert_2^2 \nonumber \\
    &\le \Vert R_E \Vert_{Lip}^2 \cdot \sup_{u, v \in V} \Vert \mb{e}_u -  \mb{e}_v\Vert_2^2 \cdot \sum_{t=1}^T \mathbb{I}(x_t \ne y_t) \nonumber \\
    &= \mathcal{O}(\text{Hamming-distance}(\mb{x}, \mb{y})).
\end{align}
The generator can easily discover better sentences on this metric by with Taylor estimator because according to \Equation{eq:UHS}, the augmented rewards come from the unit Hamming sphere.

Lipschitz and embedding constraints are implemented by spectral norm regularization \cite{yoshida2017spectral} on all layers except the embedding layer and $\ell_2$ penalty over threshold $M$ on word vectors respectively:
\begin{align}
    \mathcal{L}_{\text{reg}} = \frac{\lambda_{SN}}{2} \sum_{l=1}^L \sigma(W^l)^2 + \frac{\lambda_E}{2|V|} \sum_{v \in V} \max(\Vert \mb{e}_v \Vert^2_2 - M^2, 0),
\end{align}
where $\sigma(W^l)$ is the largest singular value of $l$th layer's weight $W^l$ and $L$ is the number of layers after the embedding layer in $D_\phi$.

As proven by~\shortcite{Arjovsky2017TowardsPM}, using objectives in \Equation{eq:D-loss}, \Equation{eq:logit_reward} without regularization is equivalent to minimizing reverse Kullback–Leibler divergence. This distance measure assigns extremely  low cost to the dropped modes. This issue may occurs when $\lambda_{SN}, \lambda_E$ are not high enough. In contrast, high $\lambda_{SN}, \lambda_E$ weaken the discriminator making it far from optimality.
As a result, we switch to the alternative in the next subsection to prevent model dropping.

\subsection{Entropy Maximization} \label{sec:entropy}
Entropy maximization overcomes the mode dropping issue by encouraging exploration and diversity \cite{liu2019onpolicy,dieng2019prescribed}. 
In previous work for reinforcement learning, a bonus term is directly added to reward such as
$(R - \lambda_{\mathcal{H}} \log \pt) \grad \log \pt$.
This term introduces extra randomness to the estimator and becomes a source of variance. Whence, we choose an alternative form considering all possible token (detailed in \Appendix{appendix_sec:entropy}) and obtain a dense objective as proposed Taylor estimator:
\begin{align}
    \grad \mathcal{H}(\pt(\cdot\mid \pre{\mb{x}})) = -\grad \sum_{v \in V} \pt(v\mid \pre{\mb{x}})\log \pt(v\mid \pre{\mb{x}}).
\end{align}
Adding the above term to \Equation{eq:taylor-estimator}, the update-formula of the generator's variable $\theta$ becomes:
\begin{align}
    \delta \theta = \hat{g}^{\text{Taylor}}_\theta + \frac{\lambda_{\mathcal{H}} }{N} \sum_{n = 1}^N \sum_{t=1}^T \grad \mathcal{H}(\pt(\cdot\mid \pre{\mb{x}^{(n)}})). \label{eq:with-entropy}
\end{align}

With the proposed components, our TaylorGAN is capable of training NLG models efficiently without a large batch size and maximum likelihood pre-training.

%% file: figures/taylor_gan.tex
\begin{figure*}[t!]
  \centering
  \begin{minipage}[b]{0.52\textwidth}
    \begin{subfigure}[t]{0.49\textwidth}
      \centering
      \includegraphics[width=\textwidth]{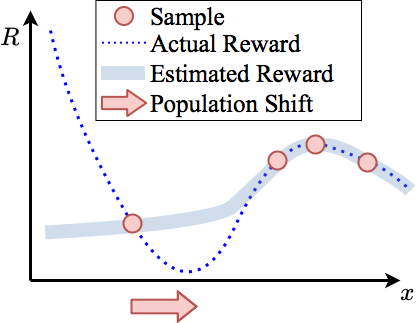}
      \caption{REINFORCE}
      \label{fig:reinforce}
    \end{subfigure}
    \begin{subfigure}[t]{0.49\textwidth}
      \includegraphics[width=\textwidth]{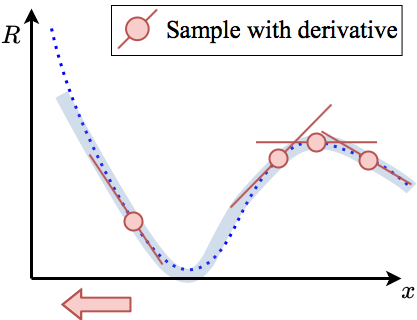}
      \caption{Taylor}
      \label{fig:taylor}
    \end{subfigure}
    \caption{Illustration of policy update estimation.}
  \end{minipage}
  \begin{minipage}[b]{0.47\textwidth}
    \centering
    \includegraphics[width=\textwidth]{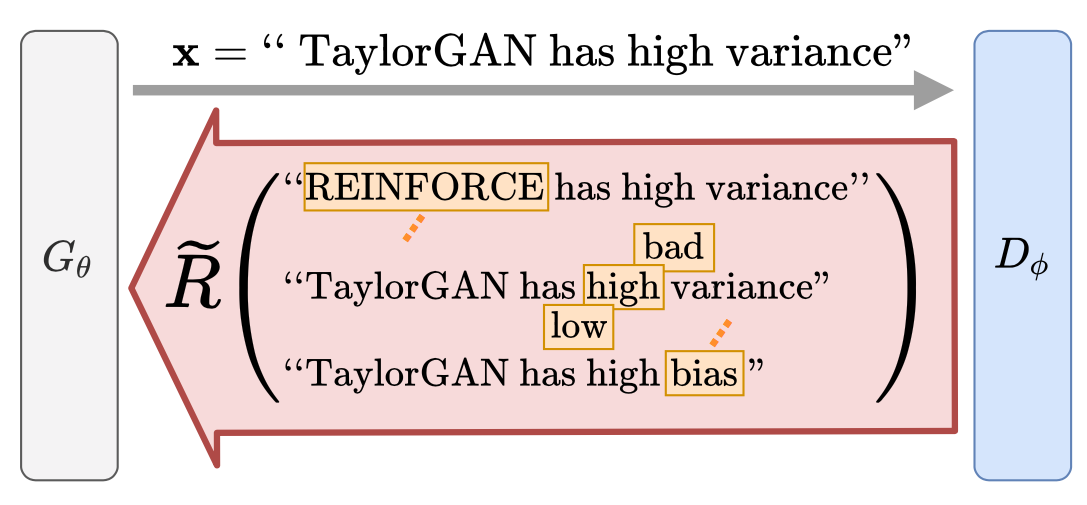}
    \caption{When a sentence $\mb{x}$ is fed to the $D_\phi$, rewards of all sentences which differ $\mb{x}$ by 1 token are provided through the backward pass.}
    \label{fig:taylor_hamming}
  \end{minipage}
\end{figure*}

%% file: sections/04-experiments.tex
In order to evaluate the effectiveness of the proposed approach, we perform a set of NLG experiments and analyze the results.

\subsection{Experimental Setup}

The dataset used in the experiments is \textbf{EMNLP 2017 News}, where sentences have a maximum length of 50 tokens and a vocabulary of 5.3k words after performing lowercase. Training and validation data consist of 269k and 10k sentences respectively.\footnote{The data is at \url{https://github.com/pclucas14/GansFallingShort/tree/master/real_data_experiments/data/news}} Additional results on COCO image caption dataset can be found in Appendix \ref{appendix_sec:coco_results}.

All models (detailed in \Appendix{appendix_sec:model}) are trained for at most 200 epochs, namely 800k training steps with the batch size $N = 64$. We save the model every epoch, and select the one with the best validation FED. Each training takes approximately 1 day on Nvidia Tesla V100 GPU.

Following \citet{caccia2018falling}, we apply temperature control by adjusting the softmax temperature parameter at test time in order to balance the trade-off between quality and diversity. Increasing the temperature makes probabilities more uniform, which leads to diverse but low-quality samples while reducing it leads to high-quality yet less diverse samples.

\subsection{Evaluation Metrics}
We evaluate our model's ability to generate realistic and diverse texts with n-gram based metric, Fréchet Embedding Distance and language model based metric.

\paragraph{N-gram Based}
BLEU \cite{papineni2002BLEU} and Self-BLEU \cite{zhu2018texygen} capture text quality and diversity respectively. Smoothing 1 proposed in \cite{chen2014systematic} is used with $\epsilon = 0.1$ for both (detailed in \Appendix{appendix_sec:bleu_smoothing}).
\begin{compactitem}
\item BLEU: a modified form of n-gram precision, measures local consistency between a set of reference and a set of candidate texts.
\item Self-BLEU: A version of BLEU in which both the reference and candidate texts are drawn from the same corpus. A high Self-BLEU means that members within the corpus are too highly correlated, indicating a lack of diversity. The size of each set is always 5k in this work.
\end{compactitem}

\paragraph{Fréchet Embedding Distance}
Fréchet Embedding Distance (FED) \cite{ScratchGAN} uses an Universal Sentence Encoder\footnote{\url{https://tfhub.dev/google/universal-sentence-encoder/2}} to measure sentimental similarity, global consistency and diversity between reference and candidate texts. 
FED is always computed with 10k against 10k samples in this work.

\paragraph{Language Model Scoring}
Following~\shortcite{rlm,caccia2018falling}, we also evaluate the quality and the diversity of generated samples by training external language model (LM) and reverse language model (RLM). We use the architecture in~\shortcite{ScratchGAN} for both.
\begin{compactitem}
\item LM:
We use a language model trained on real data to estimate the \textit{negative log likelihood per word} of generated sentences. Samples which the language model considers improper, such as ungrammatical or nonsense sentences, have a low score.
\item RLM:
If we instead train a language model on generated data while evaluating on real data, we can judge the diversity of the generated samples. Low-diversity generated data leads to an overfitted model that generalizes poorly on real data, as indicated by a low score. Models are always trained with 10k samples in this work.
\end{compactitem}

\paragraph{Perplexity}
Apart from using external language model, we can also use our generator, since it is also an autoregressive language model that produces the  probability of given sequence as shown in \Equation{eq:auto-regressive}. We can evaluate the generator's perplexity, which is the inverse of the per-word probability, for generating given real data. If there is a mode dropping problem, it will result in high perplexity.

%% file: sections/05-results.tex
In this section, we compare our results with previous approaches including MLE, LeakGAN \cite{Guo2018LongTG}, MaliGAN \cite{MaliGAN2017}, RankGAN \cite{RankGAN2017}, RelGAN \cite{nie2018relgan}, ScratchGAN \cite{ScratchGAN} and SeqGAN \cite{Yu2017SeqGANSG} \footnote{Reproduced by running the released code on \url{https://github.com/geek-ai/Texygen/} and \url{https://github.com/deepmind/deepmind-research/tree/master/scratchgan}} from different perspectives via the evaluation metrics presented in the last section. Furthermore, a quantitative study is performed to know the contribution of each technique. Samples from MLE, ScratchGAN, and TaylorGAN can also be seen in \Appendix{appendix_sec:samples}, alongside with training data.

\input{figures/bleu_lm_score_emnlp.tex}

\subsection{Quality and Diversity}

\input{tables/experiment}

On metrics of local text consistency and diversity, TaylorGAN significantly outperforms prior GAN-based models which rely heavily on pre-training except ScratchGAN. The BLEU/Self-BLEU temperature sweeps shown in \Fig{fig:val_bleu_5} for TaylorGAN indicate that TaylorGAN approaches the performance of MLE model. On the other hand, in \Fig{fig:lm_vs_rlm}, the LM scores show similar results for TaylorGAN and MLE model.

\subsection{Global Consistency}
TaylorGAN improves global consistency comparing to prior works and language model on LM score and FED score, respectively. As shown in \Table{tab:fed_lm_score}, our method reduces the gap of LM score between GAN-based method and MLE, where the latter is directly trained to minimize the negative log likelihood. Besides, in \Table{tab:fed_lm_score} we show that TaylorGAN outperforms MLE and other MLE pre-trained or large batch GAN-based methods on FED score.

\subsection{Quantitative Study}
We show the influence of gradient estimator, $\Lambda$, $\lambda_{SN}$, and $\lambda_{\mathcal{H}}$ in \Table{tab:quantitative} on validation performance, and observe that:
\begin{compactitem}
    \item The Taylor estimator outperforms Gumbel-Softmax, REINFORCE and Straight-Through baselines on FED.
    \item We argue that the inferior performance of Gumbel-Softmax is the consequence of biased and spiky distribution explained in \Section{paragraph:continuous} and the unusually high perplexity on real data, even with temperature annealing during the training phase \cite{jang2016categorical}.
    \item High $\Lambda$ results in better perplexity. We argue that it is due to the generator being forced to explore by the Taylor estimator in this case.
    \item The Taylor estimator is less reliant on discriminator constraint, showing its potential of transfer learning using a reward network trained from another source.
    \item Low $\lambda_{SN}$ leads to worse perplexity while high $\lambda_{SN}$ leads to worse FED. It supports the statement we make in \Section{sec:d-constraint}.
    \item Entropy maximization significantly reduces mode dropping indicated by much lower perplexity compared to $\lambda_{\mathcal{H}} = 0$.
\end{compactitem}

%% file: figures/bleu_lm_score_emnlp.tex
\begin{figure*}[t!]
    \centering
    \begin{subfigure}[t]{0.47\textwidth}
        \centering
        \includegraphics[width=1.0\columnwidth]{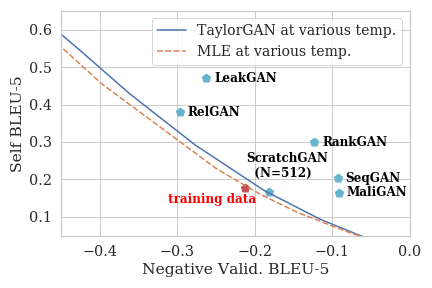}
        \caption{Negative BLEU-5 versus Self-BLEU-5.}
        \label{fig:val_bleu_5}
    \end{subfigure}
    \begin{subfigure}[t]{0.47\textwidth}
        \centering
        \includegraphics[width=\columnwidth]{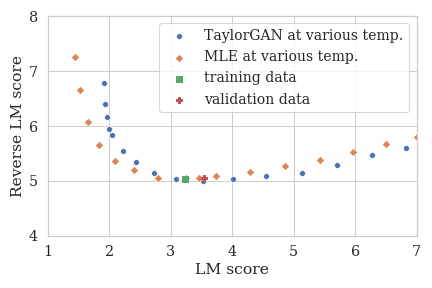}
        \caption{Language- and reverse language-model scores.}
        \label{fig:lm_vs_rlm}
    \end{subfigure}
    \caption{
        Temperature sweeps of BLEU (left) and language model scores on (right) on EMNLP 2017 News. Left and lower is better.
    }
\end{figure*}

%% file: tables/experiment.tex
\begin{table}[t!]
    \begin{minipage}[c]{.46\linewidth}
        \input{tables/comparison}
    \end{minipage}
    \hspace{3mm}
    \begin{minipage}[c]{.48\linewidth}
        \input{tables/quantitative_study}
    \end{minipage}
\end{table}

%% file: tables/comparison.tex
\caption{
    FED \& LM scores on EMNLP 2017 News. Lower is better. Batch size $N=64$ if not specified. $\dagger$ follows the settings of \protect\cite{ScratchGAN} and $\ddagger$ are pre-trained by MLE.
}
\label{tab:fed_lm_score}
\vspace{2mm}
\centering
\begin{tabular}{lrrc}
    \toprule
    \multirow{2}{*}{\bf Model}  & \multicolumn{2}{c}{\bf FED} & \bf LM \\
    \cmidrule{2-3}
    & \bf Train & \bf Val. & \bf Score\\
    \midrule
    Training data      & 0.0050 & 0.0120 & 3.22 \\
    MLE$^\dagger$      & 0.0100 & 0.0194 & \bf{3.43} \\
    \midrule
    SeqGAN$^\ddagger$  & 0.1234 & 0.1422 & 6.09 \\
    MaliGAN$^\ddagger$ & 0.1280 & 0.1504 & 6.30 \\
    RankGAN$^\ddagger$ & 0.1418 & 0.1431 & 5.76 \\
    LeakGAN$^\ddagger$ & 0.0718 & 0.0691 & 4.90 \\
    RelGAN$^\ddagger$  & 0.0462 & 0.0408 & 3.49 \\
    \midrule
    ScratchGAN         & 0.0301 & 0.0390 & 4,96 \\
    $~+N=512$         & 0.0153 & 0.0194 & 4.46 \\
    \midrule
    TaylorGAN          & 0.0105 & \bf{0.0149} & 4.02 \\
    \bottomrule
\end{tabular}

%% file: tables/quantitative_study.tex
\caption{Quantitative study on EMNLP 2017 News validation data. Lower is better.}
\label{tab:quantitative}
\vspace{2mm}
\centering
\begin{tabular}{r r r | r r}
    \toprule
    $\Lambda$ & $\lambda_{SN}$ & $\lambda_{\mathcal{H}}$ & \bf FED & \bf Perplexity \\
    \midrule
    \multicolumn{5}{l}{Gumbel-Softmax \cite{jang2016categorical}} \\
      N/A & 0.07 & 0.02 & 0.0218 & 6369 \\
    \midrule
    \multicolumn{5}{l}{REINFORCE ($\Lambda \to 0$)} \\
        & 0.07 & 0.02 & 0.0181 & 72 \\
        & \underline{0.10} & 0.02  & 0.0183 & 64 \\
        & \underline{0.03} & 0.02 & 0.0221 & 91 \\
        & \underline{0.01} & 0.02 & 0.0410 & 133 \\
        & 0.07 & \underline{0.00} & 0.0203 & 211 \\
    \midrule
    \multicolumn{5}{l}{Straight-Through ($\Lambda \to \infty$)} \\
        & 0.07 & 0.02 & 0.0449 & 116\\
    \midrule
    \multicolumn{5}{l}{Taylor ($\Lambda = 0.5$ as default)} \\
        0.50 & 0.07 & 0.02 & \bf{0.0149} & 72 \\
      \underline{0.25} & 0.07 & 0.02 & 0.0179 & 75 \\
      \underline{1.00}    & 0.07 & 0.02 & 0.0199 & \bf{60} \\
        0.50 & \underline{0.10} & 0.02 & 0.0178 & 64 \\
        0.50 & \underline{0.03} & 0.02 & 0.0183 & 91 \\
        0.50 & \underline{0.01} & 0.02 & 0.0221 & 99 \\
        0.50 & 0.07 & \underline{0.00} & 0.0200 & 310 \\
    \bottomrule
\end{tabular}

%% file: sections/06-conclusion.tex
In this work, we have presented TaylorGAN, an effective method for training an NLG model without MLE pre-training. Starting from REINFORCE, we derive a new estimator with a proper reward function that improves the sample efficiency and reduces the variance.
The benchmark experiments demonstrate that the proposed TaylorGAN improves the quality/diversity of generated texts as measured by various standard metrics.
In the future, we plan to improve our estimator for achieving low bias together with low variance, and apply it to other fields such as model-based reinforcement learning.

%% file: sections/broader_impact.tex
Improving text generation may have a wide range of beneficial impact across many domains. This includes human-machine collaboration of code, literature or even music, chatbots, question-answering systems, etc. However, this technology may be misused whether deliberately or not.

Because our model learns the underlying distribution of datasets like any other language model, it inherently risks producing biased or offensive content reflective of the training data. Studies have shown that language models could produce biased content with respect to gender, race, religion, while modeling texts from the web \citep{solaiman2019release}.
 
 Better text generation could lower costs of disinformation  campaigns and even weaken our detection ability of synthetic texts. Studies found that extremist groups can use language models for misuse specifically by finetuning the model on corresponding ideological positions \citep{solaiman2019release}.
 

%% file: sections/appendix.tex
\clearpage
\appendix
\begin{appendices}

\section{Implementation of Taylor Expansion on Unit Hamming Sphere}\label{appendix_sec:taylor-implement}

Following \Section{sec:hamming}, since all $\mb{y}$s are drawn from the unit Hamming sphere centered at $\mb{x}$, \Equation{eq:taylor} becomes a special case:
\begin{align}
    \widetilde{R}(\mb{x} \to [\mb{x}_{<t}, v, \mb{x}_{>t}]) &= R(\mb{x}) + (\mb{e}_v - \mb{e}_{x_t}) \cdot \nabla_{\mb{E}_t} R_E(\mb{E}) \bigg|_{\mb{E} = E(\mb{x})} \nonumber \\
    &= R(\mb{x}) + \mb{e}_v \cdot \nabla_{\mb{E}_t} R_E(\mb{E}) - \mb{e}_{x_t} \cdot \nabla_{\mb{E}_t} R_E(\mb{E}) \bigg|_{\mb{E} = E(\mb{x})} \nonumber \\
    &= R(\mb{x}) + \mb{e}_v^\top \nabla_{\mb{E}_t} R_E(\mb{E}) - \mb{1}_{d_E}^\top (\mb{e}_{x_t} \odot \nabla_{\mb{E}_t} R_E(\mb{E})) \bigg|_{\mb{E} = E(\mb{x})} \label{eq:taylor-hamming},
\end{align}
where $\mb{E}_t \in \mathbb{R}^{d_E}$ is the $t$th column of $\mb{E}$, $\mb{1}$ is matrix of ones, $\top$ stands for transpose, and $\odot$ stands for element-wise multiplication.

$\big\{\widetilde{R}(\mb{x} \to \mb{y})~\big|~ \mb{y} \in \bigcup_{t=1}^T \text{UHS}_t(\mb{x})\big\}$ can be computed efficiently by stacking the results of \Equation{eq:taylor-hamming} to a matrix $\widetilde{\mb{R}} \in \mathbb{R}^{|V| \times T}$ (we simplify $\nabla_{\mb{E}} R_E(\mb{E}) \bigg|_{\mb{E} = E(\mb{x})}$ as $\delta \mb{E}$ for convenience):
\begin{align}
    \widetilde{\mb{R}}(\mb{x}) &\coloneqq \begin{bmatrix} 
    \Rtilde(\mb{x} \to [\text{``apple''}, \mb{x}_{>1}]) & \dots & \Rtilde(\mb{x} \to [\mb{x}_{<T}, \text{``apple''}]) \\
    \vdots & \ddots & \vdots\\
    \Rtilde(\mb{x} \to [\text{``zebra''}, \mb{x}_{>1}]) & \dots & \Rtilde(\mb{x} \to [\mb{x}_{<T}, \text{``zebra''}])
    \end{bmatrix} \nonumber \\
    &= R(\mb{x}) \mb{1}_{|V|\times T} + W_E~ \delta\mb{E} - \mb{1}_{|V| \times d_E} (E(\mb{x}) \odot \delta\mb{E}),
\end{align}
where $W_E \coloneqq \begin{bmatrix} \mb{e}^\top_{\text{``apple''}} \\ \vdots \\\mb{e}^\top_{\text{``zebra''}} \end{bmatrix}$ is the $|V|\times d_E$ embedding matrix of all tokens and $\delta \mb{E}$ can be computed by backpropagation.

With the above operations, there is no need to explicitly collect the neighboring sequences on unit Hamming spheres.

\section{Properties of Suffix Probability Simplification} \label{appendix_sec:suffix}
Following \Section{sec:hamming} and replacing $\Rtilde$ by a function $f$ independent to suffix $\mb{x}_{>t}$, the expectation of the Taylor estimator becomes 
\begin{align}
    &\expt{\mb{x} \sim \pt} \big[\sum_{\mb{y} \in \text{UHS}_t(\mb{x})} \frac{K(y_t \mid x_t) \pt(y_t \mid \mb{x}_{<t})}{\sum_{u \in V} K(v \mid u) \pt(u \mid \mb{x}_{<t})} f(\mb{x}_{<t}, v) \grad \log \pt(v \mid \mb{x}_{<t}) \big] \nonumber\\
    =&\expt{\mb{x}_{\le t} \sim \pt} \big[\sum_{v \in V} \frac{K(v \mid x_t) \pt(v \mid \mb{x}_{<t})}{\sum_{u \in V} K(v \mid u) \pt(u \mid \mb{x}_{<t})} f(\mb{x}_{<t}, v) \grad \log \pt(v \mid \mb{x}_{<t}) \big] \nonumber\\
    =&\expt{\mb{x}_{< t} \sim \pt} \big[\sum_{x_t \in V} \pt(x_t \mid \mb{x}_{<t}) \sum_{v \in V} \frac{ K(v \mid x_t) \pt(v \mid \mb{x}_{<t})}{\sum_{u \in V} K(v \mid u) \pt(u \mid \mb{x}_{<t})} f(\mb{x}_{<t}, v) \grad \log \pt(v \mid \mb{x}_{<t})\big] \nonumber\\
    =& \expt{\mb{x}_{< t} \sim \pt} \big[\sum_{v \in V} \frac{\sum_{x_t \in V} K(v \mid x_t) \pt(x_t \mid \mb{x}_{<t})}{\sum_{u \in V} K(v \mid u) \pt(u \mid \mb{x}_{<t})} f(\mb{x}_{<t}, v) \grad \pt(v \mid \mb{x}_{<t}) \big] \nonumber\\
    =& \expt{\mb{x}_{< t} \sim \pt} \big[\sum_{v \in V} f(\mb{x}_{<t}, v) \grad \pt(v \mid \mb{x}_{<t})\big] = \grad \expt{\mb{x}_{\le t} \sim \pt}[f(\mb{x}_{<t}, x_t)].
\end{align}

Let $f(\mb{x}_{<t}, v)$ be a constant baseline $b$, we proved that replacing $\Rtilde$ with $\Rtilde - b$ introduces no bias due to the fact $\grad \expt{}[b] = \grad b = 0$.

Let $f(\mb{x}_{<t}, v)$ be the state-action value function $Q^\pt(\mb{x}_{<t}, v) \coloneqq \expt{\mb{x}_{>t} \sim \pt(\cdot \mid \pre{\mb{x}},v)}\big[R(\replace{\mb{x}}{v})\big]$, we proved that the simplified form is an unbiased estimator of $\grad \expt{\pt}[Q]$.

\section{Equivalence to REINFORCE and Straight-Through Estimators} \label{appendix_sec:RST}

Follow the discussion in \Section{paragraph:trade-off} about the special cases of bandwidth $\Lambda$. Due to the property proved in \Appendix{appendix_sec:suffix}, baseline $b$ in \Equation{eq:taylor-estimator} is eliminated in the following proofs.

\paragraph{REINFORCE}
  $\Lambda \to 0 \Rightarrow K(u \mid v) \to \ind(u = v).$ \\
  By using the identity $\Rtilde(\mb{x} \to \mb{x}) = R(\mb{x})$, the Taylor estimator in \Equation{eq:taylor-estimator} with respect to a single sample $\mb{x}$ becomes
  \begin{align}
    &\sum_{t=1}^T \sum_{\mb{y} \in \text{UHS}_t(\mb{x})} \widetilde{A}_t(\mb{x} \to \mb{y}) \grad \log\pt(y_t \mid \mb{x}_{<t}) \nonumber\\
    =& \sum_{t=1}^T \sum_{\mb{y} \in \text{UHS}_t(\mb{x})} \frac{\ind(y_t = x_t) \pt(y_t \mid \pre{\mb{x}})}{\sum_{u \in V} \ind(y_t = u) \pt(u \mid \pre{\mb{x}})}\Rtilde(\mb{x} \to \mb{y}) \grad \log\pt(y_t \mid \mb{x}_{<t}) \nonumber\\
    =& \sum_{t=1}^T \frac{\ \pt(x_t \mid \pre{\mb{x}})}{\pt(x_t \mid \pre{\mb{x}})}\Rtilde(\mb{x} \to \mb{x}) \grad \log\pt(x_t \mid \mb{x}_{<t}) \nonumber\\
    =& \sum_{t=1}^T R(\mb{x}) \grad \log \pt(x_t \mid \pre{\mb{x}}) = R(\mb{x}) \grad \log \pt(\mb{x}) = \hat{g}^{\text{REINFORCE}}_\theta(\mb{x}),
  \end{align}
  where $\hat{g}^{\text{REINFORCE}}$ is exact the same form as \Equation{eq:reinforce}.

\paragraph{Straight-Through Estimator}
  $\Lambda \to \infty \Rightarrow K(u \mid v) \to $ constant.\\
  Following \Equation{eq:st} and setting $f(\mb{h}) = g(W_E^\top \mb{h})$, where $\mb{h}$ is any $|V|$-dimensional column vector and $g: \mathbb{R}^{d_E} \to \mathbb{R}$, we have 
  \begin{align}
      \nabla_{\mb{h}} f(\mb{h}) &\bigg|_{\mb{h} = \text{one-hot}(x_t)} = \nabla_{\mb{h}}~ g(W_E^\top \mb{h}) \bigg|_{\mb{h} = \text{one-hot}(x_t)}
      = W_E \nabla_{\mb{e}} g(\mb{e}) \bigg|_{\mb{e} = W_E^\top \text{one-hot}(x) = \mb{e}_x} \nonumber\\
      \Rightarrow \nabla_{\mb{h}} f(\mb{h}) &\bigg|_{\mb{h} = \text{one-hot}(x_t)} \cdot \grad \mb{p}_\theta = \nabla_{\mb{e}} g(\mb{e}) \bigg|_{\mb{e} = \mb{e}_x} \cdot \sum_{v \in V} \mb{e}_v \grad p_\theta(v),
  \end{align}
  which implies the equivalence between \Equation{eq:st} and using the expectation of word vector as the proxy of stochastic embedding operation.
  
  Using the above property on $\pt(\cdot \mid \mb{x}_{<t})$ instead of $p_\theta$, identity $\sum_{v \in V} \grad \pt(v \mid \mb{x}_{<t}) = \grad 1 = 0$ and simplifying $\nabla_{\mb{E}_t} R_E(\mb{E}) \bigg|_{\mb{E} = E(\mb{x})}$ as $\delta \mb{E}_t$ for convenience, \Equation{eq:taylor-estimator} with respect to a single sample $\mb{x}$ and step $t$ becomes
  \begin{align}
    & \sum_{\mb{y} \in \text{UHS}_t(\mb{x})} \widetilde{A}_t(\mb{x} \to \mb{y}) \grad \log\pt(y_t \mid \mb{x}_{<t}) \nonumber\\
    =& \sum_{\mb{y} \in \text{UHS}_t(\mb{x})} \frac{ \pt(y_t \mid \pre{\mb{x}})}{\sum_{u \in V} \pt(u \mid \pre{\mb{x}})}\Rtilde(\mb{x} \to \mb{y}) \grad \log\pt(y_t \mid \mb{x}_{<t}) \nonumber\\
    =& \sum_{\mb{y} \in \text{UHS}_t(\mb{x})} \Rtilde(\mb{x} \to \mb{y}) \grad\pt(y_t \mid \mb{x}_{<t}) \nonumber\\
    =& \sum_{v \in V} (R(\mb{x}) + (\mb{e}_v - \mb{e}_{x_t}) \cdot \delta \mb{E}_t) \grad \pt(v \mid \pre{\mb{x}}) \nonumber\\
    =& ~(R(\mb{x}) - \mb{e}_{x_t} \cdot \delta\mb{E}_t) \cancelto{0}{\sum_{v \in V} \grad  \pt(v \mid \pre{\mb{x}})} + \sum_{v \in V} \mb{e}_v \cdot \delta\mb{E}_t \grad \pt(v \mid \pre{\mb{x}}) \nonumber\\
    =& ~\delta \mb{E}_t \cdot \sum_{v \in V} \mb{e}_v \grad \pt(v \mid \pre{\mb{x}}) 
    = \hat{g}^{\text{ST}}_{\theta,t}(\mb{x}).
  \end{align}
   In this case, \Equation{eq:taylor-estimator} is equivalent to using \Equation{eq:st} on the sampling operation of every step $t$.

\section{Gradient Estimator of Entropy} \label{appendix_sec:entropy}
In this section, we continue the discussion in \Section{sec:entropy} and obtain the form used in \Equation{eq:with-entropy}.
The Shannon entropy of discrete samples randomly generated by policy $\pt$ is defined as
\begin{align}
    \mathcal{H}(\pi) \coloneqq -\sum_{\mb{x} \in V^T} \pt(\mb{x}) \log \pt(\mb{x}).
\end{align}

To compute the gradient with respect to parameter $\theta$, we apply the same identity $\grad \pt = \pt \grad \log \pt$ used to derive \Equation{eq:reinforce} and get
\begin{align}
    \grad \mathcal{H}(\pt) &= -\grad \sum_{\mb{x} \in V^T} \pt(\mb{x}) \log \pt(\mb{x}) \nonumber\\
    &= -\sum_{\mb{x} \in V^T} \log \pt(\mb{x}) \grad \pt(\mb{x}) + \cancel{\pt(\mb{x})} \frac{\grad \pt(\mb{x})}{\cancel{\pt(\mb{x})}} \nonumber\\
    &= -\sum_{\mb{x} \in V^T} \log \pt(\mb{x}) \grad \pt(\mb{x}) - \cancelto{0}{\sum_{\mb{x} \in V^T} \grad \pt(\mb{x})} \nonumber\\
    &= \expt{\mb{x} \sim \pt}[-\log \pt(\mb{x}) \grad \log \pt(\mb{x})],
\end{align}

which has a similar form as REINFORCE with $R_{\mathcal{H}}(\mb{x}) \coloneqq -\log \pt(\mb{x})$ as objective. Thus, $R_{\mathcal{H}}$ can be added to the original reward to encourage diversity.

Furthermore, due to \Equation{eq:auto-regressive}, $R_{\mathcal{H}}(\mb{x})$ can be written as $\sum_{t=1}^T r_t(\mb{x})$ with $r_t(\mb{x}) \coloneqq -\log \pt(x_t \mid \mb{x}_{<t})$, which provides per-word rewards. By applying causality and discounting the future reward with a factor $\gamma$, we rewrite the update signal of $\theta$ from the entropy bonus as
\begin{align}
    \hat{g}^{\mathcal{H}, \gamma}_\theta \coloneqq \expt{ \mb{x} \sim \pt}\big[\sum_{t=1}^T \grad \log \pt(x_t \mid \mb{x}_{<t}) \sum_{t'=t}^T \gamma^{t' - t} r_{t'}(\mb{x}) \big].
\end{align}

In this work, we use $\gamma \to 0$ for a greedy reward, which ignores the suffix $\mb{x}_{>t}$ and allows us to write a dense form by expanding all possible next token $x_t$:
\begin{align}
    \hat{g}^{\mathcal{H}, 0}_\theta &= \expt{\mb{x} \sim \pt}\big[\sum_{t=1}^T \grad \log \pt(x_t \mid \mb{x}_{<t})~ r_t(\mb{x}) \big] \nonumber\\
    &= -\expt{\mb{x} \sim \pt}\big[\sum_{t=1}^T \grad \log \pt(x_t \mid \mb{x}_{<t}) \log \pt(x_t \mid \mb{x}_{<t})) \big] \nonumber\\
    &= -\sum_{t=1}^T \expt{\mb{x}_{<t} \sim \pt}\big[\sum_{v \in V} \pt(v \mid \mb{x}_{<t}) \grad \log \pt(v \mid \mb{x}_{<t}) \log \pt(v \mid \mb{x}_{<t})) \big] \nonumber\\
    &= -\sum_{t=1}^T \expt{\mb{x}_{<t} \sim \pt}\big[\sum_{v \in V} \grad \pt(v \mid \mb{x}_{<t}) \log \pt(v \mid \mb{x}_{<t})) \big] \nonumber\\
    &= -\expt{\mb{x} \sim \pt}\big[\sum_{t=1}^T \sum_{v \in V} \grad \pt(v \mid \mb{x}_{<t}) \log \pt(v \mid \mb{x}_{<t})) \big] = \expt{\mb{x} \sim \pt}\big[\sum_{t=1}^T \grad \mathcal{H}(\pt(\cdot \mid \mb{x}_{<t}))\big],
\end{align}
where $\mathcal{H}(\pt(\cdot \mid \mb{x}_{<t}))$ is the entropy of next-token distribution.
This biased but dense form works well in practice. 

\section{Model Architecture} \label{appendix_sec:model}

\input{tables/model_architecture.tex}

\subsection{Discriminator}
The architecture of the discriminator is shown in \Table{tab:D-architecture}. Exponential Linear Units \cite{clevert2015fast}, whose continuous $\nth{1}$ derivative and sparse $\nth{2}$ derivative satisfy the condition of Taylor's theorem in \Equation{eq:taylor}, and reduce the norm of Hessian matrix, is used as all non-linear transformations.

To perform consistent convolution on sentences of different lengths, we also implement a special masking mechanism. For all convolution and pooling layers, we mask out input features where the receptive field is out-of-bounds with respect to the unpadded input sentence. As shown in \Fig{fig:masking}, the same padding is performed as if $T_{\text{input}}$ had been 3, even though it is actually 5 (not including the leftmost $\mb{0}$ added by original implementation of same padding) with 2 PAD tokens . 

\subsection{Generator}
The generator we use is an autoregressive probabilistic model with GRU \cite{chung2014empirical} shown in \Fig{fig:rnn-workflow}.
In the linear layer, the output of GRU is projected to the dimension of word vectors and then multiplied by the transpose of embedding matrix \cite{press2016using} to obtain the softmax logits.
The start-of-sentence token is fed to the embedding layer at $t=1$. The vocabulary also contains an end-of-sentence token. If the generator outputs this token at any time step $t_{\text{EOS}}$, the sentence ends and its length is set to $t_{\text{EOS}}$.

\section{Training Details}
The discriminator and generator are updated with a 1:1 ratio and both trained with Adam \cite{kingma2014adam} of learning rate $ = 10^{-4}, \beta_1=0.5$ and $\beta_2=0.999$. Gradients are clipped by maximum global norm $=10$. $\Lambda = 0.5, \lambda_{SN} = 0.07, \lambda_{E} = 0.2, M = 1$ and $\lambda_{\mathcal{H}} = 0.02$ if not specified.
Baseline $b$ is assigned as the exponential moving average of $R$ with decay rate $=0.9$.
All word vectors are initialized with pre-trained fastText embeddings \cite{bojanowski2017enriching}.

\paragraph{BLEU smoothing}
\label{appendix_sec:bleu_smoothing}

One of the issue with BLEU is that in the case that a higher order n-gram precision of a sentence is 0, then the BLEU score will be 0, resulting in severely underestimation. This is due to the fact that BLEU is calculated by the geometric mean of precision. To solve this, we replaced the precision by the smoothed version as follows:
\begin{align}
    \text{smoothed-precision}_n \coloneqq \max\left(\text{precision}_n, \frac{\epsilon}{\text{Count(n-grams)}}\right).
\end{align}
We've picked the smoothing factor $\epsilon = 0.1$ as proposed by \citet{chen2014systematic}.

\section{COCO Results} \label{appendix_sec:coco_results}

\begin{table*}[t!]
    \begin{minipage}[t]{.47\linewidth}
        \input{figures/bleu_score_coco}
    \end{minipage}
    \begin{minipage}[t]{.51\linewidth}
        \input{figures/sentence_length_statistics}
    \end{minipage}
\end{table*}

Sentences in the COCO dataset have a maximum length of 24 tokens and a vocabulary of 4.6k words after performing lowercase. Training and validation data both consist of 10k sentences.\footnote{The preprocessed COCO dataset is available at \url{https://github.com/pclucas14/GansFallingShort/tree/master/real_data_experiments/data/coco}}.
The BLEU/Self-BLEU temperature sweeps is shown in \Fig{fig:val_bleu_5_coco}.

\section{Generated Samples} \label{appendix_sec:samples}
\input{tables/samples.tex}
Samples of EMNLP2017 News from MLE, ScratchGAN, and TaylorGAN can be seen in \Table{tab:samples}, alongside with training data. The senetence length statistics is shown in \Fig{fig:length}.

\input{tables/coco_samples}
Samples of COCO from MLE, LeakGAN, MaliGAN, RankGAN, RelGAN, SeqGAN, TextGAN, and TaylorGAN can be seen in \Table{tab:samples-coco}, alongside with training data.

\end{appendices}

%% file: tables/model_architecture.tex
\begin{table}[t]
  \begin{minipage}[b]{0.22\textwidth}
    \begin{tabular}{l}
    	\toprule
    	\midrule
     	Embedding \\
    	\midrule
        Conv3-512 ELU \\
        Conv4-512 ELU \\
        Mean Pool2 \\
        \midrule
        Conv3-1024 ELU \\
        Conv4-1024 ELU \\
        Global Mean Pool \\
        \midrule
        Dense 1024 ELU \\
        Dense 1 Sigmoid \\
        \midrule
    	\bottomrule
    \end{tabular}
    \vspace{8mm}
    \centering
    \caption{Discriminator.}
    \label{tab:D-architecture}
  \end{minipage}
  \begin{minipage}[b]{0.45\textwidth}
    \includegraphics[width=\columnwidth]{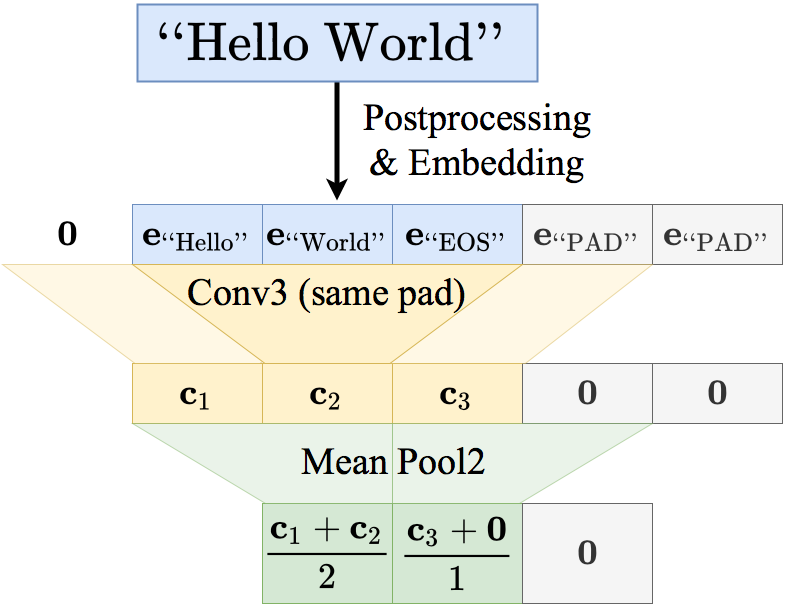}
    \vspace{1mm}
    \centering
    \captionof{figure}{Masking mechanism.}
    \label{fig:masking}
  \end{minipage}
  \begin{minipage}[b]{0.32\textwidth}
    \includegraphics[width=\columnwidth]{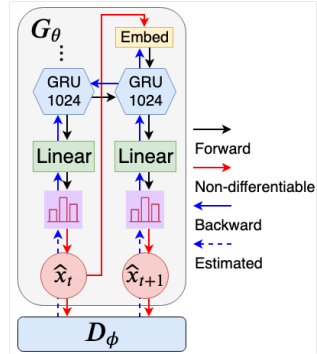}
    \vspace{2mm}
    \captionof{figure}{Generator.}
    \label{fig:rnn-workflow}
 \end{minipage}
\end{table}

%% file: figures/bleu_score_coco.tex
    \centering
    \includegraphics[width=1.0\columnwidth]{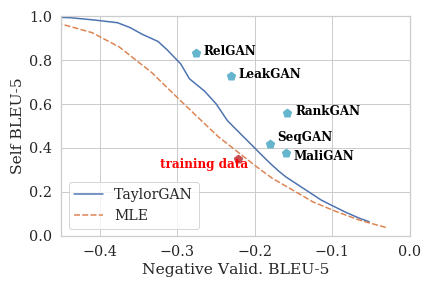}
    \captionof{figure}{
        Temperature sweeps on COCO. \\ Left and lower is better.
    }
    \label{fig:val_bleu_5_coco}

%% file: figures/sentence_length_statistics.tex
    \centering
    \includegraphics[width=0.9\textwidth]{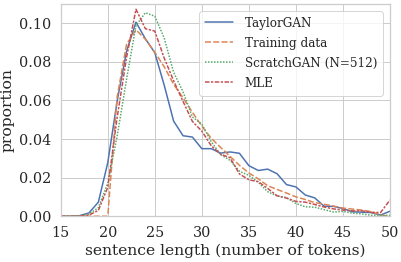}
    \vspace{1mm}
    \captionof{figure}{Sentence length statistics on EMNLP2017 News.}
    \label{fig:length}

%% file: tables/samples.tex
\begin{table}[t!]
  \caption{
    Training data, MLE, ScratchGAN, TaylorGAN samples of EMNLP 2017 News with temperature $= 1.0$. $N$ is the batch size used to train each model.
  }
  \label{tab:samples}
  \centering
  \begin{tabular}{p{0.94\linewidth}}
    \toprule
    \textbf{Training Data} \\ 
    \midrule
    - the u . s . has struggled to find an effective ground force to take on isis in syria , where president obama has ruled out a u . s . ground combat role . \\
    - it is a happy and kind world that we live in on this show and that is where i hope we can live in real life . \\
    - while men and women may end up earning roughly the same amount in the same jobs , men are more likely to end up in higher - paying roles in the tech industry . \\
    - but while they were beaten by a better side , the tie did reveal what i think has been city ' s biggest problem this season : they have lost the ability to score against good teams . \\
    - according to facebook ' s policies , accounts can be suspended if law enforcement believe individuals are at risk of harm . \\
    \midrule
    \textbf{MLE} ($N=64$) \\ 
    \midrule
    - a one - game event and a limited number this dropped by 38 per cent to find out what we could want , to be as much done as we finish as the games . \\
    - the uk government got on the deal when it came to other eu countries that allowed workplace checks for 26 . \\
    - so women i didn ’ t want my parents to work hard with me at the time of the task of making it presents . \\
    - black voters feel there may not be enough choice between former president george w . bush and i have to go along with problems and support the people that are in a swing state for president . \\
    - let ’ s catch him down his back phone one morning john f kennedy , where he is in a us meeting group with supporters of love not being successful in the fields . \\
    \midrule
    \textbf{ScratchGAN} ($N=512$) \\ 
    \midrule
    - they were together to be twice a week of a junior doctor who work pushed their home but they can necessarily have to have a size to go one of strikes on . \\
    - democrats cannot overcome a support of mr . barack obama , and obama has an influence we have other aspects of the most powerful president and fought supporting the gop nominee . \\         
    - the researchers thought that women looked older for more than boys having been used in terms of the sexual assault from the swedish victims . \\
    - " since i get engaged with the woman you share all wonderful and got out of my foot traveling to the main daily network , everything told me he their sister were dead and showing what sort of position there and , " he said . \\
    \midrule
    \textbf{TaylorGAN} ($N=64$) \\ 
    \midrule
    - " well , i don ' t think it will ever be a good opportunity , " murray told the press . \\
    - " however , they do not know how much the information needed to be discovered , " the company said in a statement . \\
    - i ' m so much younger than i wanted to , and it didn ' t sense so much for me . \\
    - " i think it was good enough that she was ready to take the time and just let her run this , " she said . \\
    - when you see people coming together , that ' s pretty different from what we ' ve taught in america . \\
    \bottomrule
  \end{tabular}
\end{table}

%% file: tables/coco_samples.tex
\begin{table}[t!]
  \caption{
    Training data, MLE, LeakGAN, MaliGAN, RankGAN, RelGAN, SeqGAN, TaylorGAN samples of COCO image caption with temperature $= 1.0$.
  }
  \label{tab:samples-coco}
  \centering
  \begin{tabular}{p{0.94\linewidth}}
    \toprule
    \textbf{Training Data} \\ 
    \midrule
    - an industrial kitchen with white walls and stainless steel counters . \\
    - a team of chefs work together to prepare a meal . \\
    - a jar filled with liquid sits on a wood surface . \\
    - a picture taken from the driver seat of car  at a stop sign . \\
    - a black and white cat sits near a window looking outside . \\
    \midrule
    \textbf{MLE} \\ 
    \midrule
    - here are motorcyclists on an trailer in front of folded . \\
    - a man with a bandana is flipping \\
    - a cat is resting on and pans and buttons of a bar and claw foot steel refrigerator has graffiti on the floor in it . \\
    - a kitchen in a bench next to a tree in it \\
    - a pretty orange parked on a table . \\
    \midrule
    \textbf{LeakGAN} \\ 
    \midrule
    - a group of people outside a jet airplane . \\
    - a bathroom with a glass shower , sink , and tub . \\
    - a black cat sitting nearby . \\
    - a modern kitchen features microwave and refrigerator . \\
    - a door showing the mirror in the reflection . \\
    \midrule
    \textbf{MaliGAN} \\ 
    \midrule
    - a man standing in a kitchen has an island with the stems \\
    - a man standing in a giraffe cake . \\
    - a car seat showing some break in a kitchen  counter . \\
    - a group of people that have ski down men standing next to a white accents . \\
    - a toilet site in top in a neighborhood . \\
    \midrule
    \textbf{RankGAN} \\ 
    \midrule
    - a bathroom filled with sink , mirrors and white tile door . \\
    - a bathroom with a large door with a blue window . \\
    - a nice bathroom that is very clean table a yellow shirt and white umbrella at a stop . \\
    - a bathroom with a shiny grey scheme a bathroom . \\
    - a table with a flower dress is perched on it 's at a field . \\
    \midrule
    \textbf{RelGAN} \\ 
    \midrule
    - a bathroom with a white toilet next to a sink and a toilet paper roller in a bathroom . \\
    - a red kitchen with glass doors open and red cupboards . \\
    - a young boy cuts a cake designed to look like - a skateboard on the stove . \\ 
    - a cat resting on top of a toilet seat in a bathroom . \\
    - dirt bikers racing on a runway under the cabinet . \\
    \midrule
    \textbf{SeqGAN} \\
    \midrule
    - a template example of an airplane in a blue field \\
    - small child looking at the crucifix on an umbrella on a plate . \\
    - an older woman is sitting on a leash . \\
    - the night of men is sitting up a very tall building . \\
    - a close up of four planes on the street together . \\
    \midrule
    \textbf{TaylorGAN} \\ 
    \midrule
    - a group of people sitting around a police motorcycle . \\
    - two people are near some airplanes are next . \\
    - several white fixtures of a modern bathroom sink . \\
    - an image of a woman in the kitchen . \\
    - a herd of sheep grazing in the middle of a runway . \\
    \bottomrule
  \end{tabular}
\end{table}

%% file: main.bbl
\begin{thebibliography}{}

\bibitem[\protect\citeauthoryear{Arjovsky and
  Bottou}{2017}]{Arjovsky2017TowardsPM}
Mart{\'i}n Arjovsky and L{\'e}on Bottou.
\newblock Towards principled methods for training generative adversarial
  networks.
\newblock {\em ArXiv}, abs/1701.04862, 2017.

\bibitem[\protect\citeauthoryear{Bengio \bgroup \em et al.\egroup
  }{2013}]{Bengio2013EstimatingOP}
Yoshua Bengio, Nicholas L{\'e}onard, and Aaron~C. Courville.
\newblock Estimating or propagating gradients through stochastic neurons for
  conditional computation.
\newblock {\em CoRR}, abs/1308.3432, 2013.

\bibitem[\protect\citeauthoryear{Bojanowski \bgroup \em et al.\egroup
  }{2017}]{bojanowski2017enriching}
Piotr Bojanowski, Edouard Grave, Armand Joulin, and Tomas Mikolov.
\newblock Enriching word vectors with subword information.
\newblock {\em Transactions of the Association for Computational Linguistics},
  5:135--146, 2017.

\bibitem[\protect\citeauthoryear{Caccia \bgroup \em et al.\egroup
  }{2018}]{caccia2018falling}
Massimo Caccia, Lucas Caccia, William Fedus, Hugo Larochelle, Joelle Pineau,
  and Laurent Charlin.
\newblock Language gans falling short.
\newblock {\em CoRR}, abs/1811.02549, 2018.

\bibitem[\protect\citeauthoryear{Che \bgroup \em et al.\egroup
  }{2017}]{MaliGAN2017}
Tong Che, Yanran Li, Ruixiang Zhang, R.~Devon Hjelm, Wenjie Li, Yangqiu Song,
  and Yoshua Bengio.
\newblock Maximum-likelihood augmented discrete generative adversarial
  networks.
\newblock {\em CoRR}, abs/1702.07983, 2017.

\bibitem[\protect\citeauthoryear{Chen and Cherry}{2014}]{chen2014systematic}
Boxing Chen and Colin Cherry.
\newblock A systematic comparison of smoothing techniques for sentence-level
  bleu.
\newblock In {\em Proceedings of the Ninth Workshop on Statistical Machine
  Translation}, pages 362--367, 2014.

\bibitem[\protect\citeauthoryear{Chung \bgroup \em et al.\egroup
  }{2014}]{chung2014empirical}
Junyoung Chung, Caglar Gulcehre, KyungHyun Cho, and Yoshua Bengio.
\newblock Empirical evaluation of gated recurrent neural networks on sequence
  modeling.
\newblock {\em arXiv preprint arXiv:1412.3555}, 2014.

\bibitem[\protect\citeauthoryear{Clevert \bgroup \em et al.\egroup
  }{2015}]{clevert2015fast}
Djork-Arn{\'e} Clevert, Thomas Unterthiner, and Sepp Hochreiter.
\newblock Fast and accurate deep network learning by exponential linear units
  (elus).
\newblock {\em arXiv preprint arXiv:1511.07289}, 2015.

\bibitem[\protect\citeauthoryear{de Masson~d'Autume \bgroup \em et al.\egroup
  }{2019}]{ScratchGAN}
Cyprien de~Masson~d'Autume, Shakir Mohamed, Mihaela Rosca, and Jack Rae.
\newblock Training language gans from scratch.
\newblock In H.~Wallach, H.~Larochelle, A.~Beygelzimer, F.~d'~Alch\'{e}-Buc,
  E.~Fox, and R.~Garnett, editors, {\em Advances in Neural Information
  Processing Systems 32}, pages 4300--4311. Curran Associates, Inc., 2019.

\bibitem[\protect\citeauthoryear{Dieng \bgroup \em et al.\egroup
  }{2019}]{dieng2019prescribed}
Adji~B Dieng, Francisco~JR Ruiz, David~M Blei, and Michalis~K Titsias.
\newblock Prescribed generative adversarial networks.
\newblock {\em arXiv preprint arXiv:1910.04302}, 2019.

\bibitem[\protect\citeauthoryear{Fedus \bgroup \em et al.\egroup
  }{2018}]{fedus2018maskgan}
William Fedus, Ian Goodfellow, and Andrew~M Dai.
\newblock Maskgan: better text generation via filling in the\_.
\newblock {\em arXiv preprint arXiv:1801.07736}, 2018.

\bibitem[\protect\citeauthoryear{Fu}{2006}]{fu2006gradient}
Michael~C Fu.
\newblock Gradient estimation.
\newblock {\em Handbooks in operations research and management science},
  13:575--616, 2006.

\bibitem[\protect\citeauthoryear{Glynn}{1990}]{glynn1990likelihood}
Peter~W Glynn.
\newblock Likelihood ratio gradient estimation for stochastic systems.
\newblock {\em Communications of the ACM}, 33(10):75--84, 1990.

\bibitem[\protect\citeauthoryear{Goodfellow \bgroup \em et al.\egroup
  }{2014}]{NIPS2014_5423}
Ian Goodfellow, Jean Pouget-Abadie, Mehdi Mirza, Bing Xu, David Warde-Farley,
  Sherjil Ozair, Aaron Courville, and Yoshua Bengio.
\newblock Generative adversarial nets.
\newblock In Z.~Ghahramani, M.~Welling, C.~Cortes, N.~D. Lawrence, and K.~Q.
  Weinberger, editors, {\em Advances in Neural Information Processing Systems
  27}, pages 2672--2680. Curran Associates, Inc., 2014.

\bibitem[\protect\citeauthoryear{Grathwohl \bgroup \em et al.\egroup
  }{2017}]{RELAX2017}
Will Grathwohl, Dami Choi, Yuhuai Wu, Geoffrey Roeder, and David~K. Duvenaud.
\newblock Backpropagation through the void: Optimizing control variates for
  black-box gradient estimation.
\newblock {\em CoRR}, abs/1711.00123, 2017.

\bibitem[\protect\citeauthoryear{Gu \bgroup \em et al.\egroup
  }{2016}]{Muprop2016}
Shixiang Gu, Sergey Levine, Ilya Sutskever, and Andriy Mnih.
\newblock Muprop: Unbiased backpropagation for stochastic neural networks.
\newblock In {\em 4th International Conference on Learning Representations
  (ICLR)}, May 2016.

\bibitem[\protect\citeauthoryear{Guo \bgroup \em et al.\egroup
  }{2018}]{Guo2018LongTG}
Jiaxian Guo, Sidi Lu, Han Cai, Weinan Zhang, Yong Yu, and Jun Wang.
\newblock Long text generation via adversarial training with leaked
  information.
\newblock In {\em AAAI}, 2018.

\bibitem[\protect\citeauthoryear{Jang \bgroup \em et al.\egroup
  }{2016}]{jang2016categorical}
Eric Jang, Shixiang Gu, and Ben Poole.
\newblock Categorical reparameterization with gumbel-softmax.
\newblock {\em CoRR}, abs/1611.01144, 2016.

\bibitem[\protect\citeauthoryear{Karras \bgroup \em et al.\egroup
  }{2019}]{karras2019style}
Tero Karras, Samuli Laine, and Timo Aila.
\newblock A style-based generator architecture for generative adversarial
  networks.
\newblock In {\em Proceedings of the IEEE Conference on Computer Vision and
  Pattern Recognition}, pages 4401--4410, 2019.

\bibitem[\protect\citeauthoryear{Kingma and Ba}{2014}]{kingma2014adam}
Diederik~P Kingma and Jimmy Ba.
\newblock Adam: A method for stochastic optimization.
\newblock {\em arXiv preprint arXiv:1412.6980}, 2014.

\bibitem[\protect\citeauthoryear{Kusner and
  Hern{\'a}ndez-Lobato}{2016}]{Kusner2016GANSFS}
Matt~J. Kusner and Jos{\'e}~Miguel Hern{\'a}ndez-Lobato.
\newblock Gans for sequences of discrete elements with the gumbel-softmax
  distribution.
\newblock {\em ArXiv}, abs/1611.04051, 2016.

\bibitem[\protect\citeauthoryear{Lin \bgroup \em et al.\egroup
  }{2017}]{RankGAN2017}
Kevin Lin, Dianqi Li, Xiaodong He, Zhengyou Zhang, and Ming-ting Sun.
\newblock Adversarial ranking for language generation.
\newblock In I.~Guyon, U.~V. Luxburg, S.~Bengio, H.~Wallach, R.~Fergus,
  S.~Vishwanathan, and R.~Garnett, editors, {\em Advances in Neural Information
  Processing Systems 30}, pages 3155--3165. Curran Associates, Inc., 2017.

\bibitem[\protect\citeauthoryear{Liu \bgroup \em et al.\egroup
  }{2019}]{liu2019onpolicy}
Jingbin Liu, Xinyang Gu, Dexiang Zhang, and Shuai Liu.
\newblock On-policy reinforcement learning with entropy regularization.
\newblock {\em arXiv preprint arXiv:1912.01557}, 2019.

\bibitem[\protect\citeauthoryear{Magnus and Neudecker}{1979}]{10.2307/2958818}
Jan~R. Magnus and H.~Neudecker.
\newblock The commutation matrix: Some properties and applications.
\newblock {\em The Annals of Statistics}, 7(2):381--394, 1979.

\bibitem[\protect\citeauthoryear{Nie \bgroup \em et al.\egroup
  }{2019}]{nie2018relgan}
Weili Nie, Nina Narodytska, and Ankit Patel.
\newblock Rel{GAN}: Relational generative adversarial networks for text
  generation.
\newblock In {\em International Conference on Learning Representations}, 2019.

\bibitem[\protect\citeauthoryear{Papineni \bgroup \em et al.\egroup
  }{2002}]{papineni2002BLEU}
Kishore Papineni, Salim Roukos, Todd Ward, and Wei-Jing Zhu.
\newblock Bleu: a method for automatic evaluation of machine translation.
\newblock In {\em Proceedings of the 40th annual meeting of the Association for
  Computational Linguistics}, pages 311--318, 2002.

\bibitem[\protect\citeauthoryear{Press and Wolf}{2016}]{press2016using}
Ofir Press and Lior Wolf.
\newblock Using the output embedding to improve language models.
\newblock {\em arXiv preprint arXiv:1608.05859}, 2016.

\bibitem[\protect\citeauthoryear{{Radford} \bgroup \em et al.\egroup
  }{2015}]{2015arXiv151106434R}
Alec {Radford}, Luke {Metz}, and Soumith {Chintala}.
\newblock {Unsupervised Representation Learning with Deep Convolutional
  Generative Adversarial Networks}.
\newblock {\em arXiv e-prints}, page arXiv:1511.06434, Nov 2015.

\bibitem[\protect\citeauthoryear{Solaiman \bgroup \em et al.\egroup
  }{2019}]{solaiman2019release}
Irene Solaiman, Miles Brundage, Jack Clark, Amanda Askell, Ariel Herbert-Voss,
  Jeff Wu, Alec Radford, Gretchen Krueger, Jong~Wook Kim, Sarah Kreps, et~al.
\newblock Release strategies and the social impacts of language models.
\newblock {\em arXiv preprint arXiv:1908.09203}, 2019.

\bibitem[\protect\citeauthoryear{Williams}{1992}]{williams1992simple}
Ronald~J Williams.
\newblock Simple statistical gradient-following algorithms for connectionist
  reinforcement learning.
\newblock {\em Machine learning}, 8(3-4):229--256, 1992.

\bibitem[\protect\citeauthoryear{Yoshida and
  Miyato}{2017}]{yoshida2017spectral}
Yuichi Yoshida and Takeru Miyato.
\newblock Spectral norm regularization for improving the generalizability of
  deep learning.
\newblock {\em arXiv preprint arXiv:1705.10941}, 2017.

\bibitem[\protect\citeauthoryear{Yu \bgroup \em et al.\egroup
  }{2017}]{Yu2017SeqGANSG}
Lantao Yu, Weinan Zhang, Jun Wang, and Yong Yu.
\newblock Seqgan: Sequence generative adversarial nets with policy gradient.
\newblock In {\em AAAI}, 2017.

\bibitem[\protect\citeauthoryear{Zhang \bgroup \em et al.\egroup
  }{2017}]{zhang2017adversarial}
Yizhe Zhang, Zhe Gan, Kai Fan, Zhi Chen, Ricardo Henao, Dinghan Shen, and
  Lawrence Carin.
\newblock Adversarial feature matching for text generation.
\newblock In {\em Proceedings of the 34th International Conference on Machine
  Learning-Volume 70}, pages 4006--4015. JMLR. org, 2017.

\bibitem[\protect\citeauthoryear{Zhao \bgroup \em et al.\egroup }{2017}]{rlm}
Junbo~Jake Zhao, Yoon Kim, Kelly Zhang, Alexander~M. Rush, and Yann LeCun.
\newblock Adversarially regularized autoencoders for generating discrete
  structures.
\newblock {\em CoRR}, abs/1706.04223, 2017.

\bibitem[\protect\citeauthoryear{Zhou \bgroup \em et al.\egroup
  }{2019}]{zhou2019lipschitz}
Zhiming Zhou, Jiadong Liang, Yuxuan Song, Lantao Yu, Hongwei Wang, Weinan
  Zhang, Yong Yu, and Zhihua Zhang.
\newblock Lipschitz generative adversarial nets.
\newblock {\em arXiv preprint arXiv:1902.05687}, 2019.

\bibitem[\protect\citeauthoryear{Zhu \bgroup \em et al.\egroup
  }{2018}]{zhu2018texygen}
Yaoming Zhu, Sidi Lu, Lei Zheng, Jiaxian Guo, Weinan Zhang, Jun Wang, and Yong
  Yu.
\newblock Texygen: A benchmarking platform for text generation models.
\newblock {\em SIGIR}, 2018.

\end{thebibliography}
